\documentclass{article}

\usepackage{microtype}
\usepackage{graphicx}
\usepackage{subcaption}
\usepackage{booktabs} 
\usepackage{booktabs} 
\usepackage{multirow} 
\usepackage{booktabs}   
\usepackage{makecell}   
\usepackage{multirow}   
\usepackage{graphicx}   
\usepackage[table]{xcolor} 
\usepackage{pifont}
\usepackage[most]{tcolorbox}

\newcommand{\cmark}{\ding{51}} 
\newcommand{\xmark}{\ding{55}} 
\usepackage{hyperref}

\usepackage[preprint]{icml2026}

\usepackage{amsmath}
\usepackage{amssymb}
\usepackage{mathtools}
\usepackage{amsthm}

\theoremstyle{plain}

\theoremstyle{definition}

\theoremstyle{remark}

\usepackage[textsize=tiny]{todonotes}

\begin{document}

\twocolumn[

  \icmltitle{Latent Context Compilation: Distilling Long Context into Compact Portable Memory}



  \icmlsetsymbol{equal}{\textdagger}
  
  \begin{icmlauthorlist}
    \icmlauthor{Zeju Li}{yyy,comp}
    \icmlauthor{Yizhou Zhou}{equal,comp}
    \icmlauthor{Qiang Xu}{equal,yyy}

  \end{icmlauthorlist}

  \icmlaffiliation{yyy}{The Chinese University of Hong Kong}
  \icmlaffiliation{comp}{Bytedance}

  \icmlcorrespondingauthor{Zeju Li}{zjli24@cse.cuhk.edu.hk}
  \icmlcorrespondingauthor{Yizhou Zhou}{zhouyizhou@bytedance.com}
  \icmlcorrespondingauthor{Qiang Xu}{qxu@cse.cuhk.edu.hk}


  \vskip 0.3in
]



\printAffiliationsAndNotice{\textdagger  Corresponding Author}  

\begin{abstract}

Efficient long-context LLM deployment is stalled by a dichotomy between amortized compression, which struggles with out-of-distribution generalization, and Test-Time Training, which incurs prohibitive synthetic data costs and requires modifying model weights, creating stateful parameters that complicate concurrent serving.
We propose \textbf{Latent Context Compilation}, a framework that fundamentally shifts context processing from adaptation to compilation. 
By utilizing a disposable LoRA module as a compiler, we distill long contexts into compact \emph{buffer tokens}---stateless, portable memory artifacts that are plug-and-play compatible with frozen base models. 
Crucially, we introduce a self-aligned optimization strategy
 that eliminates the need for synthetic context-relevant QA pairs. 
By regularizing context reconstruction task with context-agnostic random queries, we force compressed tokens to reside within the model's existing instruction-following manifold. Experiments with Llama-3.1-8B demonstrate that Latent Context Compilation preserves fine-grained details and reasoning capabilities where prior methods falter, effectively decoupling memory density from model parameters even at a 16$\times$ compression ratio.
\end{abstract}

\section{Introduction}


While LLMs now support million-token windows, the quadratic cost of attention and prohibitive KV cache footprints create a ``Context Bottleneck" that renders scalable deployment cost-prohibitive. Consequently, High-Fidelity Context Compression has become imperative to reduce input length without compromising semantic integrity.

Current approaches generally fall into two paradigms, each facing a fundamental trade-off. 
The first, Amortized Compression (or General Context Compression), learns a fixed policy to compress inputs. 
Methods like LLMLingua \cite{pan2024llmlingua}, 500xCompressor \cite{li-etal-2025-500xcompressor}, and ICAE \cite{ge2023context} are efficient but suffer from a generalization gap: a compressor trained on a pre-defined corpus inevitably struggles to capture the fine-grained nuances of out-of-distribution (OOD) contexts at test time. They often sacrifice detail for brevity, making them unsuitable for precision-critical tasks.

To bridge this gap, recent research has pivoted towards Instance-Specific Optimization, leveraging the emerging paradigm of ``Test-Time Compute" to adapt models dynamically. 
This landscape is generally divided into two streams, yet neither fully resolves the efficiency-deployment dilemma. 

\begin{itemize}
    \item \emph{Test-Time Adaptation (TTA)} \cite{hu2025test, niu2024testtimemodeladaptationforward}, optimizes the model to align with the test data distribution. While effective for mitigating domain shifts, TTA primarily targets prediction robustness rather than dimensionality reduction; the raw context is still required as input, leaving the prohibitive computational overhead unresolved. 
    \item \emph{Test-Time Training (TTT)} \cite{wang2024greater,cao2025infiniteiclbreakinglimitcontext,muhtar2024streamadapterefficienttesttime}, attempts to address this by ``memorizing" specific context content directly into model parameters. However, this introduces significant \emph{architectural friction}. By encoding memory into weights, these methods transform the LLM from a stateless engine into a stateful entity. While these weights are technically swappable, managing concurrent requests with different parameter states precludes standard KV-caching optimizations and complicates high-throughput serving. Furthermore, this parameter coupling risks degrading general reasoning capabilities due to overfitting on the specific context.
\end{itemize}


\begin{table*}[h]
\centering
\caption{Comparison of our Latent Context Compilation with other methods on training stage and inference stage.}
\label{tab:method_comparison_v4}
\resizebox{\textwidth}{!}{%
\begin{tabular}{lccccc}
\toprule
\multirow{2}{*}{\textbf{Method Hierarchy}} & \multicolumn{3}{c}{\textbf{Training Stage}} & \multicolumn{2}{c}{\textbf{Inference Stage}} \\
\cmidrule(lr){2-4} \cmidrule(lr){5-6}
 & \textbf{Pretraining} & \textbf{Context-relevant Queries} & \textbf{Optimization Target} & \textbf{Inference Dependency} & \textbf{Frozen Parameter} \\
\midrule

\textbf{General Compression} & & & & & \\

\quad LLMLingua2 \cite{pan2024llmlingua} 
 & \textbf{\cmark}
 & \textbf{\xmark}
 & \multirow{3}{*}{\makecell{Global Encoder\\$\theta_{enc}$ (Fixed)}} 
 & \multirow{3}{*}{\makecell{Memory Slots /\\Soft Prompts}} 
 & \multirow{3}{*}{\cmark} \\

\quad AutoCompressor \cite{chevalier2023adapting} 
 & \textbf{\cmark}  
 & \textbf{\xmark} 
 & & & \\ 

\quad GIST \cite{mu2023learning} 
 & \textbf{\cmark}  
 & \textbf{\xmark}  
 & & & \\
\midrule

\textbf{Test Time Adaption} & & & & & \\

\quad TLM \cite{hu2025test} 
 & \textbf{\xmark}   
 & \textbf{\cmark}  
 & \multirow{3}{*}{\makecell{Model Weights\\$\Delta \theta$ (Distribution)}} 
 & \multirow{3}{*}{\makecell{LoRA Weights +\\Raw Context}} 
 & \multirow{3}{*}{\xmark} \\

\quad  SIFT \cite{hubotter2024efficiently}
 & \textbf{\xmark}  
 & \textbf{\cmark}  
 & & & \\

\quad TTT-NN \cite{hardt2023test}
 & \textbf{\xmark}  
 & \textbf{\cmark}  
 & & & \\
\midrule

\textbf{Test Time Training} & & & & & \\

\quad Temp-LoRA \cite{wang2024greater} 
 & \textbf{\xmark}   
 & \textbf{\xmark}  
 & \multirow{3}{*}{\makecell{Model Weights\\$\Delta \theta$ (Instance-Specific)}} 
 & \multirow{3}{*}{\makecell{LoRA Weights}} 
 & \multirow{3}{*}{\xmark} \\

\quad StreamAdapter \cite{muhtar2024streamadapterefficienttesttime} 
 & \textbf{\xmark}  
 & \textbf{\cmark}  
 & & & \\

\quad InfiniteICL \cite{cao2025infiniteiclbreakinglimitcontext}
 & \textbf{\xmark}  
 & \textbf{\cmark}  
 & & & \\
\midrule

\rowcolor{blue!5} 
\textbf{Latent Context Compilation} 
 & \textbf{\xmark}  
 & \textbf{\xmark}  
 & \textbf{\makecell{Buffer Tokens}}
 & \textbf{\makecell{Buffer Tokens\\(KV Cache Only)}} 
 & \textbf{\cmark} \\
\bottomrule
\end{tabular}%
    }
\vspace{-10pt}
\end{table*}

In this paper, we propose \textbf{Latent Context Compilation}, a novel framework that fundamentally shifts context processing from \textit{adaptation} to \textit{compilation}. 
Synthesizing the instance-specificity of TTT with the deployment efficiency of standard caching, we leverage a disposable LoRA module as a compiler to distill the context into a compact set of \textbf{Buffer Tokens} (Input KV cache). 
Crucially, this results in \textit{Data Portability} rather than weight portability: the LoRA is discarded after optimization, leaving a portable memory artifact that can be plugged into the frozen base model as standard input. 
To ensure this compilation is robust without ground-truth supervision, we introduce a self-aligned optimization strategy: we optimize for Context Reconstruction Task to ensure high fidelity, while simultaneously regularizing the model with context-agnostic random queries to maintain its general instruction-following manifold. 
A detailed comparison of methodological differences is presented in Table \ref{tab:method_comparison_v4}, highlighting how Latent Context Compilation overcomes the stateful limitations of TTT and the generalization gap of Amortized Compression.


To validate our approach, we employ Llama-3.1-8B-Instruct \cite{grattafiori2024llama} across a diverse suite of benchmarks. 
On context-specific QA and summarization tasks, \textbf{Latent Context Compilation} significantly outperforms baselines, confirming that our buffer tokens successfully encode both the fine-grained details and global semantics required for high-fidelity retrieval. 
Crucially, on general reasoning benchmarks, we demonstrate that the model maintains performance parity with the original base model. 
This proves that our context self-aligned training successfully prevents the catastrophic forgetting common in prior distillation works, preserving the model's fundamental capabilities while compressing memory.

Our contributions are summarized as follows: 
\begin{itemize}
    
    \item We introduce \textbf{Latent Context Compilation}, a framework that uses a disposable LoRA to compile contexts into portable Buffer Tokens. This decouples memory from model weights, overcoming the stateful serving bottlenecks of Test-Time Training.
    \item We propose a self-aligned optimization strategy that combines context reconstruction task with agnostic queries regularization. This prevents manifold collapse and ensures high-fidelity compression without the cost or bias of synthetic data.
    \item Experiments demonstrate that Latent Context Compilation achieves a 16$\times$ - 32$\times$ compression ratio with significantly higher retention than extractive and weight-based baselines, while fully preserving general reasoning capabilities.
\end{itemize}

\section{Related Work}

\subsection{General Context Compression}

Current strategies for long-context efficiency primarily rely on learning a general-purpose compression policy over large-scale datasets, generally falling into extractive pruning or amortized soft compression categories. Extractive pruning strategies, such as LLMLingua \cite{jiang2023llmlingua}, LLMLingua-2 \cite{pan2024llmlingua}, KV cache compression \cite{liu2025chunkkv, devoto2025expectedattention} and Selective Context \cite{li2023compressingcontext}, operate by employing budget controllers or distilled token classifiers to explicitly identify and remove tokens deemed redundant based on information-theoretic metrics. However, physically severing tokens from the sequence risks disrupting long-range dependencies and diminishing context integrity \cite{li2024snapkv, zhang2023h2o}, particularly in tasks requiring fine-grained reasoning.

Alternatively, amortized soft compression approaches focus on condensing text into compact latent representations or memory slots to preserve semantic information. This paradigm encompasses autoencoder-based models like ICAE \cite{ge2023context} and AutoCompressors \cite{chevalier2023adapting}, which leverage massive pre-training on autoencoding objectives to encode long contexts into summary vectors. Other works explore soft prompt learning, such as Gist Tokens \cite{mu2023learning, deng2025silver} and contrastive conditioning \cite{wingate2022promptcompressioncontrastiveconditioning}, or introduce specialized compression modules, including Activation Beacon \cite{zhang2024longactivationbeacon}, retrieval-augmented compressors like RECOMP \cite{xu2023recomp}, and lightweight adapters for streaming contexts \cite{kim2024compressedcontextmemoryonline, liu2025contextcascadecompressionexploring}. While these amortized methods offer efficient inference, they fundamentally suffer from the \textit{generalization gap}: they rely on a fixed compression function learned from a pre-defined training distribution. Consequently, they often generalize poorly to out-of-distribution (OOD) texts and require expensive pre-training to cover diverse domains.




\subsection{Test-Time Adaptation} 
Test-Time Adaptation (TTA) focuses on adapting the model to specific test instances to mitigate distribution shifts, primarily aiming to improve prediction accuracy rather than achieving dimensionality reduction. Standard approaches, such as TENT \cite{wang2020tent} and TLM \cite{hu2025test}, employ gradient-based updates to minimize prediction entropy or perplexity on the unlabeled test input. To reduce the computational overhead of backpropagation, recent works have explored gradient-free paradigms. For instance, {Forward-Optimization Adaptation} \cite{niu2024testtimemodeladaptationforward} introduces a method to adapt prompts and shift activations using only forward passes, enabling adaptation on edge devices or quantized models without modifying weights. Other strategies, such as TTT-NN \cite{hardt2023test} and SIFT \cite{hubotter2024efficiently}, retrieve relevant examples from a training bank to refine the model's inference state.

However, since TTA primarily focuses on optimizing the model to align with the test domain distribution (domain adaptation) rather than \textit{internalizing} the information content itself, the raw context remains a necessary input during inference. Consequently, these methods fail to alleviate the computational burden of processing long contexts.

\subsection{Test-Time Training}
A distinct line of work leverages Test-Time Training specifically for context storage, with the goal of ``memorizing" the context into model parameters so that the original text can be discarded. Prominent methods such as Temp-LoRA \cite{wang2024greater} and StreamAdapter \cite{muhtar2024streamadapterefficienttesttime} train a temporary LoRA module on the context (often augmented with synthetic QA pairs) to internalize the information. Similarly, InfiniteICL \cite{cao2025infiniteiclbreakinglimitcontext} draws parallels between LLM parameters and long-term memory, proposing to transform temporary context knowledge into permanent parameter updates to theoretically support infinite context integration.

However, distilling a specific context into model weights inherently biases the model, often causing catastrophic forgetting of general instruction-following capabilities. Furthermore, since the compressed state exists as non-transferable model parameters rather than data, context switching becomes computationally expensive and structurally incompatible with standard KV-cache pipelines.


\section{Method}
\begin{figure*}
    \centering
    \includegraphics[width=0.9\linewidth]{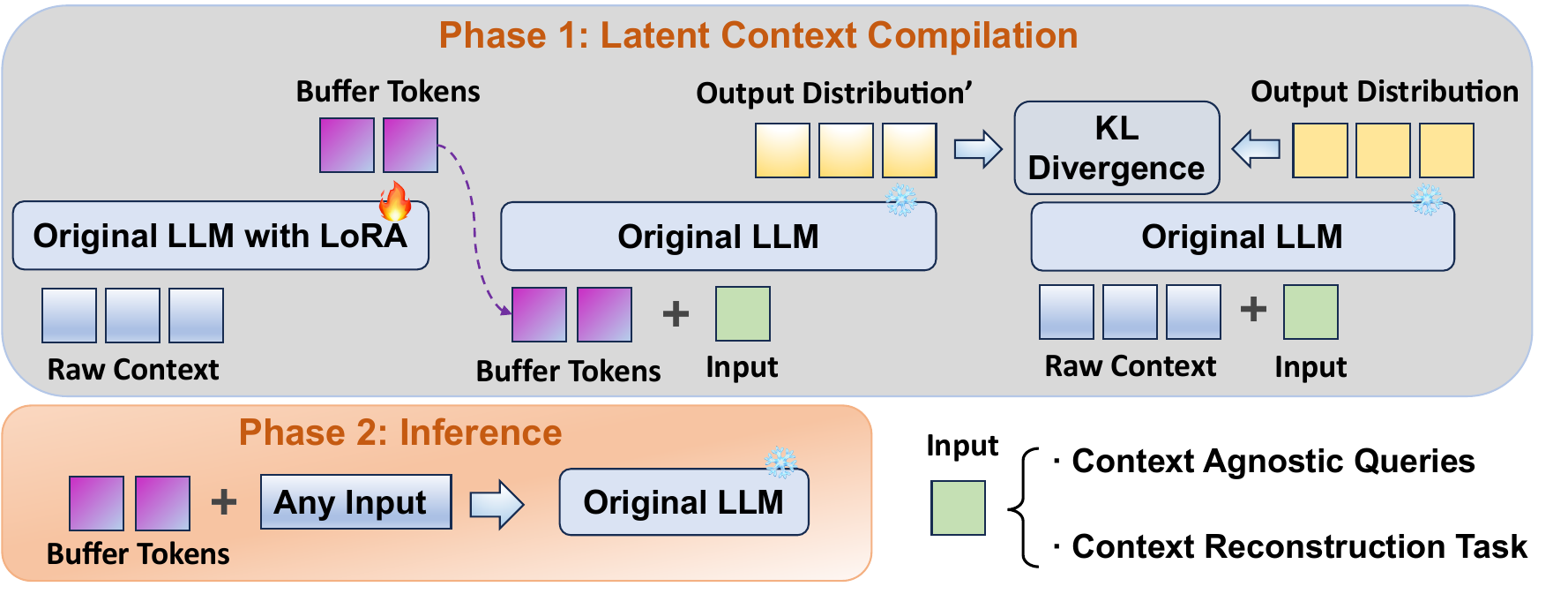}
    \caption{Overview of Latent Context Compilation. During Phase 1, we distill the raw context into compact Buffer Tokens using a disposable LoRA that forces information flow through the buffer. The model minimizes KL divergence against a full-context teacher. During Phase 2, the LoRA module is discarded to ensure portability. The resulting Buffer Tokens are retained as a standard KV cache, allowing the frozen LLM to perform high-fidelity inference on new queries with zero additional parameters.}
    \label{fig:1}
\end{figure*}

We propose Latent Context Compilation, a framework designed to distill a long context $C$ into a compact sequence of buffer tokens $T_{buf}$ at inference time. Our design philosophy centers on portability and data independence: the resulting compressed state must be compatible with the frozen base model without requiring additional weights, and the optimization must proceed without ground-truth supervision. The overview of the framework is illustrated in Figure \ref{fig:1}.

\subsection{Theoretical Formulation}
The core objective of high-fidelity compression is to ensure that the compressed state $T_{buf}$ is functionally equivalent to the original context $C$ under the operation of the language model. Specifically, to maintain this equivalence, we must guarantee that for any arbitrary input, the corresponding output distribution remains consistent. We first define the context compression process as a learnable mapping function $\mathcal{F}_{\phi}$:
\begin{equation}
\label{eq:compression}
T_{buf} = \mathcal{F}_{\phi}(C)
\end{equation}
where $\phi$ denotes the parameters of the trainable LoRA adapters injected into the model. Formally, for any arbitrary query $x$, we seek to minimize the divergence between the output distribution (logits) induced by the original context and buffer tokens:
\begin{equation}
\label{eq:objective}
\min_{T_{buf}} \mathcal{L} = \mathbb{E}_{x \sim \mathcal{D}_{\text{query}}} \left[ D_{\text{KL}} \left( P_{\theta}(\cdot | C, x) \parallel P_{\theta}(\cdot | T_{buf}, x) \right) \right]
\end{equation}
where $\theta$ denotes the frozen base model parameters. 
Rather than requiring equivalence across all random non-semantic inputs, we constrain optimization to semantic queries $x \sim \mathcal{D}_{\text{query}}$, prioritizing the preservation of reasoning capabilities over robustness to irrelevant noise.
Since the true distribution of user queries $\mathcal{D}_{\text{query}}$ is unknown at test time, we approximate this expectation using a constructed surrogate dataset $\mathcal{D}_{\text{surrogate}}$, described in Section \ref{sec:data}.

\subsection{Compressive Bottleneck Architecture}To enforce the compression constraint, we introduce a Bottleneck Architecture governed by a strict causal mask. We prepend $K$ learnable buffer tokens, $T_{buf} = \{b_1, ..., b_K\}$, to the input sequence. The attention mechanism $A(Q, K, V)$ is modified via a custom mask matrix $M$:
\begin{itemize}
    \item \textbf{Buffer Tokens ($T_{buf}$):} Can attend to the raw context $C$ and prior buffer tokens. This enables the aggregation of semantic information from $C$ into the buffer state.
    \item \textbf{Context Agnostic Query \& Response ($x, y$):} Can \textit{only} attend to the Buffer Tokens $T_{buf}$ and themselves. Crucially, they are attention-isolated from the raw context $C$ (i.e., $M_{ij} = -\infty$ for $i \in \{x, y\}, j \in C$).
    \end{itemize}
This strategy creates a rigorous information bottleneck: the conditional probability $P(y|C, x)$ collapses to $P(y|T_{buf}, x)$, forcing the optimization process to encode all necessary information into the buffer embeddings to satisfy the objective in Eq. \ref{eq:objective}.

\subsection{Disposable LoRA as a Compression Catalyst}
Optimizing the input embeddings of $T_{buf}$ directly via backpropagation is insufficient for capturing deep semantic dependencies due to the limited expressivity of the input space. To bridge this gap, we introduce a Disposable LoRA module $\phi$ applied to the attention layers.Gradient Isolation Strategy. Unlike standard TTT methods that treat LoRA weights as the storage medium (creating a portability bottleneck), we treat LoRA solely as a temporary ``catalyst" for compression. We enforce a strict separation of computational graphs:

\textbf{Compression Stage ($C \to T_{buf}$):}  As formalized in Equation \ref{eq:compression}, the context is processed through the LoRA-augmented model ($\theta + \phi$). Gradients flow through $\phi$ to optimize the projection of $C$ into $T_{buf}$.

\textbf{Generation Stage ($T_{buf} \to y$):} The LoRA module is disabled. The model relies \textit{solely} on the optimized $T_{buf}$ and the frozen base weights $\theta$ to generate the output logits.

This Gradient Isolation ensures buffer tokens are aligned with the \textit{frozen} manifold of original model. Once optimization concludes, $\phi$ is discarded, leaving a standard KV cache that is plug-and-play compatible with original model.

\subsection{Optimization Strategy}
\label{sec:data}

Directly optimizing Eq. \ref{eq:objective} is challenging due to the absence of ground-truth QA pairs. To address this, we employ a \textbf{self-aligned optimization strategy}, constructing a surrogate dataset $\mathcal{D}_{\text{surrogate}}$ composed of two tasks. We term this ``self-aligned'' as it leverages the base model's intrinsic manifold to regularize the compression, ensuring compatibility without external supervision.

\textbf{Manifold Regularization via Context-Agnostic Queries.} To ensure the learned buffer tokens remain compatible with the base model's generation mechanics, we sample Context-Agnostic Queries ($Q_{rnd}$) from generic instruction datasets (e.g., Alpaca \cite{alpaca}). 
This design avoids the prohibitive latency of test-time generation and prevents the model from overfitting to the inductive biases inherent in synthetic context relevant data, which often degrade generalizability.
We utilize the frozen base model as a teacher to generate soft labels:
\begin{equation}
\mathcal{L}_{reg} = D_{KL}(P_{\theta}(y|C, Q_{rnd}) \parallel P_{\theta+\phi}(y|T_{buf}, Q_{rnd}))
\end{equation}
Although $C$ is semantically irrelevant to $Q_{rnd}$, enforcing distributional alignment on these queries constrains the buffer tokens to reside within the valid instruction-following manifold of the LLM. This prevents the compression module from overfitting to a specific pattern and collapsing the model's ability to understand general queries.

\textbf{Fidelity Enhancement via Context Reconstruction Task.}
However, relying solely on unrelated queries provides insufficient supervision for fine-grained detail retention. To explicitly force the buffer tokens to encode the exact informational content of $C$, we introduce a Context Reconstruction task. We construct samples where the instruction is \textit{``Please repeat the context"} and the target is $C$ itself:
\begin{equation}
\mathcal{L}_{recon} = D_{KL}(P_{\theta}(C|C) \parallel P_{\theta+\phi}(C|T_{buf}))
\end{equation}
Minimizing this reconstruction loss maximizes the mutual information between the original context and the compressed buffer. As demonstrated in our ablation studies (Sec. \ref{sec:data_ablation}), combining this explicit memory objective with manifold regularization is strictly necessary to achieve high-fidelity compression performance.

\subsection{Inference Strategy}
Upon convergence, we perform a final forward pass to compute the KV cache for $T_{buf}$. The LoRA module $\phi$ is then discarded. The resulting compressed KV cache is a standalone, portable memory artifact that can be plugged into any instance of the original frozen model to service infinite future queries.

\begin{table*}[h]
\centering
\caption{\textbf{Main Results.}  Comparison of Context retention ratio and Performance across Different Datasets for Specific and General Tasks.}
\label{tab:merged_score_comparison}
\resizebox{\textwidth}{!}{%
\begin{tabular}{@{}lcccccccc@{}}
\toprule
\multirow{2}{*}{\textbf{Method}} & \textbf{Context Retention} & \multicolumn{5}{c}{\textbf{Context Specific Tasks $\uparrow$}} & \multicolumn{2}{c}{\textbf{General Tasks $\uparrow$}} \\
\cmidrule(lr){3-7} \cmidrule(lr){8-9}
 & \textbf{Ratio} & \textbf{SQuAD} & \textbf{Fiction} & \textbf{CoQA} & \textbf{BookSum} & \textbf{XSum} & \textbf{GPQA} & \textbf{Alpaca} \\
\midrule
\rowcolor{gray!10} \multicolumn{9}{l}{\textit{Bounds}} \\
Base Model w/ Context (Upper) & 100\% & 3.18 & 4.36 & 2.89 & 4.90 & 3.86 & 0.89 & 2.88 \\
Base Model w/o Context (Lower) & 0 & 0.80 & 0.40 & 0.30 & 0.00 & 0.00 & \textcolor{gray}{1.24} & \textcolor{gray}{3.69} \\
\midrule
\rowcolor{gray!10} \multicolumn{9}{l}{\textit{Compression \& Pruning Baselines}} \\
LLMLingua-2 & 20\% & 0.55 & 1.20 & 0.65 & 0.76 & 0.96 & \textbf{1.20} & 2.97 \\
ChunkKV & 20\% & 0.57 & 1.09 & 0.93 & 2.20 & 0.50 & 0.52 & 1.83 \\
Expected Attention & 20\% & {2.94} & {2.75} & {2.70} & {2.40} & {3.22} & 0.73 & 2.15 \\
\midrule
\rowcolor{gray!10} \multicolumn{9}{l}{\textit{Test-Time Adaptation (TTA)}} \\
TLM (Original) & 0 & 1.05 & 0.38 & 0.23 & 0.00 & 0.00 & 1.03 & {3.08} \\
TLM (Ours Data) & 0 & 1.07 & 0.46 & 0.31 & 0.12 & 0.09 & 1.10 & \textbf{3.09} \\
\midrule
\rowcolor{gray!10} \multicolumn{9}{l}{\textit{Test-Time Training (TTT)}} \\
Temp-LoRA & 0 & 0.50 & 0.93 & 0.42 & 2.00 & 1.00 & 0.51 & 1.97 \\
InfiniteICL & 0 & 0.45 & 1.10 & 0.81 & 0.20 & 0.40 & 0.47 & 1.60 \\
\midrule
\rowcolor{blue!5} \textbf{Latent Context Compilation (16x)} & \textbf{6.25\%} & \textbf{3.01} & \textbf{4.08} & \textbf{3.29} & \textbf{4.10} & \textbf{3.30} & 0.80 & 2.73 \\
\bottomrule
\end{tabular}%
}
\vspace{-10pt}
\end{table*}

\section{Experiments}


\subsection{Experimental Setup}


\paragraph{Data Construction \& Hyperparameters.}
Adhering to our Data-Free protocol, we construct the training set $\mathcal{D}_{surrogate}$ without ground-truth supervision. We combine Context Reconstruction samples (target: raw context) with Context-Agnostic Queries randomly sampled from the Alpaca training dataset \cite{alpaca}. Supervision is provided exclusively by the frozen teacher via the KL-divergence objective. Regarding data scale, we adopt a balanced configuration of 2,000 reconstruction samples and 2,000 unrelated queries for standard benchmarks. 
The details of data scaling is analyzed in Section \ref{sec:data_ablation}.

\paragraph{Model \& Training Implementation.}
We utilize Llama-3.1-8B-Instruct as our backbone. The compression framework consists of a disposable LoRA module applied to attention layers ($r=8, \alpha=16$) and a set of learnable buffer tokens. For main results, we standardize the compression ratio to 16$\times$, distilling the long context into compact tokens. We validate the robustness of this ratio (exploring 2$\times$ to 32$\times$) in Section \ref{sec:compression_scaling}. 
We optimize the LoRA module using AdamW with a learning rate of $2\times 10^{-5}$ and BF16 precision. Training runs for 45 epochs with an effective batch size of 8 (per-device batch size 1 with 8 gradient accumulation steps) to ensure convergence.

\subsection{Baselines \& Evaluation Protocols}

\paragraph{Baselines.} We evaluate our Latent Context Compilation against representative methods spanning three efficient inference paradigms. To benchmark extractive and soft compression approaches, we include \textbf{LLMLingua-2} \cite{pan2024llmlingua} (token pruning), \textbf{ChunkKV} \cite{liu2025chunkkv}, and \textbf{Expected Attention} \cite{devoto2025expectedattention}. In the realm of weight-based Test-Time Training (TTT), we compare against \textbf{Temp-LoRA} \cite{wang2024greater} and \textbf{InfiniteICL} \cite{cao2025infiniteiclbreakinglimitcontext}, both of which attempt to encode context directly into temporary model parameters. We also assess Test-Time Adaptation (TTA) via \textbf{TLM} \cite{hu2025test}. \textit{Crucially, to ensure a fair comparison of information retention, we modify the standard TLM inference protocol by discarding the raw context during generation, thereby forcing the model to rely solely on the adapted weights.} Finally, we contextualize these results within two theoretical bounds: {Base Model w/ Context} (Upper Bound) representing ideal retention, and {Base Model w/o Context} (Lower Bound) reflecting pure hallucination capabilities.

\paragraph{Benchmarks.} Our evaluation strategy is strictly categorized into Context-Specific Tasks for retention and General Tasks for safety. For the Context-Specific cluster, which includes \textbf{SQuAD 2.0} \cite{rajpurkar2018know}, \textbf{CoQA} \cite{reddy2019coqa}, \textbf{BookSum} \cite{kryscinski2021booksum}, and \textbf{XSum} \cite{Narayan2018DontGM}, we randomly sample 5 distinct long contexts from each dataset to conduct a focused assessment. Additionally, we introduce a synthetic \textbf{Fictional Story} dataset generated by AI, specifically designed to test fine-grained recall; each story is paired with 15 questions spanning easy, medium, and hard difficulty levels. For General Tasks, specifically \textbf{GPQA} \cite{rein2023gpqagraduatelevelgoogleproofqa} and \textbf{Alpaca Eval}, we sample 100 instances for each to verify robustness against catastrophic forgetting. 

\paragraph{Metrics.} To ensure a nuanced and consistent quality assessment across these diverse tasks, we employ an LLM-as-a-Judge framework using GPT-4o. This judge evaluates model responses on a multi-dimensional scale from 0 to 5, prioritizing accuracy, coherence, and faithfulness over rigid lexical overlap metrics. We define specific qualitative anchors to standardize evaluation: scores of \textbf{4--5} denote high-fidelity responses that are factually complete and fully consistent with the context; scores of \textbf{3--4} indicate mostly correct answers with minor omissions or slight inconsistencies; while scores of \textbf{2--3} reflect partially correct responses containing noticeable hallucinations or significant information gaps. Scores below 2 indicate failure to retrieve relevant information. The case studies are illustrated in Appendix \ref{app:judge_cases}.

\subsection{Main Results: Fidelity vs. Robustness}

Table \ref{tab:merged_score_comparison} presents a comprehensive evaluation of our method against general compression, TTA and TTT baselines. We structure the analysis along two critical dimensions: (1) \textbf{Context-Specific Tasks}, which assess the fidelity of the compressed representations in retaining fine-grained information; and (2) \textbf{General Tasks}, which evaluate robustness by querying the model on context agnostic benchmarks while conditioned on the compressed context, thereby measuring potential capability drift.

\textbf{Superior Fidelity via Latent Context Compilation.} On context-specific tasks, Latent Context Compilation demonstrates a decisive advantage over extractive compression paradigms despite operating at a significantly higher compression rate. While baselines like LLMLingua-2 and ChunkKV retain 20\% of the original tokens (5$\times$ compression), they suffer severe degradation due to the physical removal of information. In contrast, our method retains only 6.25\% of the tokens (16$\times$ compression) yet achieves performance parity with the full-context Upper Bound across all five benchmarks. This confirms that our Latent Context Compilation distills semantic dependencies far more effectively than rigid token pruning. Notably, on CoQA, our method even {surpasses} the upper bound (3.29 vs. 2.89). We hypothesize that our manifold regularization implicitly acts as a denoising filter, suppressing colloquial distractions in the raw context to produce a more refined retrieval signal.

We further compare against Test-Time Adaptation (TTA) and Training (TTT) paradigms, where the context is distilled entirely into model weights (0\% token retention). For TLM, we evaluate it using both standard benchmarks and our context-agnostic data; in both settings, it fails to retain sufficient information when the raw context is discarded, significantly underperforming our method. Similarly, Weight-based TTT methods (Temp-LoRA, InfiniteICL) exhibit a systematic collapse near the ``No Context" lower bound. This is particularly revealing for InfiniteICL, which explicitly synthesizes context-relevant QA pairs to enforce memorization. Its failure points to a fundamental {``Storage Medium Mismatch"}: model parameters are optimized for slowly-evolving semantic knowledge, not for the rapid, one-shot storage of transient episodic memory. Forcing gradients to encode high-frequency details creates an unstable optimization landscape, rendering weights a far less effective storage medium than the input activations (buffer tokens) utilized by our method.

\textbf{Robustness via Manifold Alignment.} On general tasks (GPQA, Alpaca), our method exhibits remarkable stability, maintaining performance metrics that are statistically on par with the ``Base Model w/ Context" upper bound. This stands in sharp contrast to weight-based approaches, which suffer from catastrophic forgetting (e.g., Temp-LoRA degrades to 0.51 on GPQA). This parity proves that our context self-aligned optimization strategy successfully confines compression artifacts to the buffer tokens, preserving the base model's original reasoning engines and instruction-following manifold intact.


\subsection{Mechanism Analysis: Necessity of Gradient Isolation} \label{sec:lora_ablation}

\begin{table}[htbp]
\centering
\caption{Results of the LoRA ablation study. We evaluate the impact of gradient isolation on CoQA dataset and context-agnostic Alpaca benchmark.}
\label{tab:lora}
\begin{tabular}{@{}lccccc@{}}
\toprule
\textbf{Method} & \textbf{COQA} & \textbf{Alpaca} \\
\midrule
Base Model w/ Context & 2.89 & 2.88 \\
{Inference w/ LoRA} & 2.46 & 2.63    \\
\textbf{Latent Context Compilation} & \textbf{3.29} & \textbf{2.73} \\

\bottomrule
\end{tabular}
\end{table}

To validate the architectural decision of freezing the base model, we compare our method against a ``Coupled" variant in generation stage. In this setting, the LoRA module remains active and trainable during both the compression and generation phases, allowing the model weights to co-adapt with the buffer tokens. Table \ref{tab:lora} reports the results.

Counter-intuitively, the ``Coupled" variant significantly underperforms ours on the CoQA benchmark ($2.46$ vs. $3.29$), despite possessing more trainable parameters. We hypothesize that allowing model weights to shift creates an optimization shortcut, where the model overfits LoRA parameters to specific surface patterns rather than compressing semantic information into the buffer tokens. In contrast, our Gradient Isolation enforces a ``hard constraint" by freezing generation weights, compelling the optimization to channel all information strictly into the buffer tokens and align them with the base model's existing manifold. This not only yields superior information density—evidenced by our method surpassing the full-context upper bound—but also prevents the capability drift observed in the coupled approach (which drops to $2.63$ on Alpaca). Furthermore, omitting the LoRA module from the coupled variant causes performance to collapse to near zero, confirming that its buffer tokens become deeply entangled with specific weight shifts; Latent Context Compilation avoids this entirely, yielding a standalone, plug-and-play memory artifact.

\begin{figure}
    \centering
    \includegraphics[width=1\linewidth]{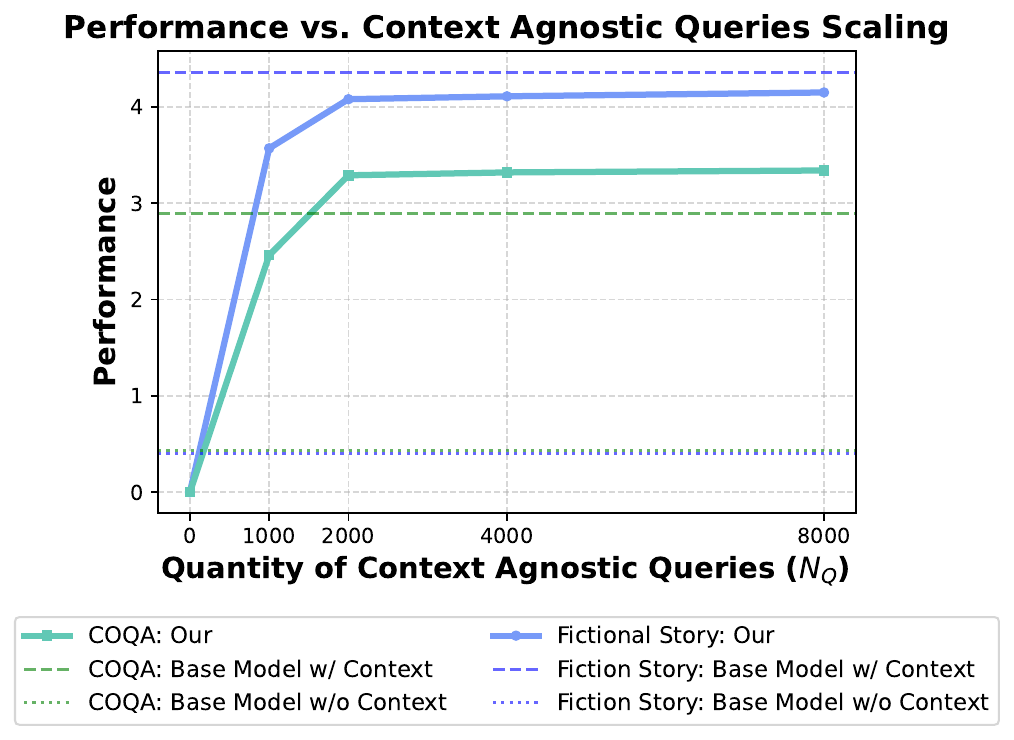}
    \caption{Scaling of Manifold Regularization. Performance trajectory on Fictional Story and CoQA datasets as the quantity of context agnostic queries ($N_{Q}$) increases. }
    \label{fig:uq_scaling}
\end{figure}


\begin{figure}
    \centering
    \includegraphics[width=1\linewidth]{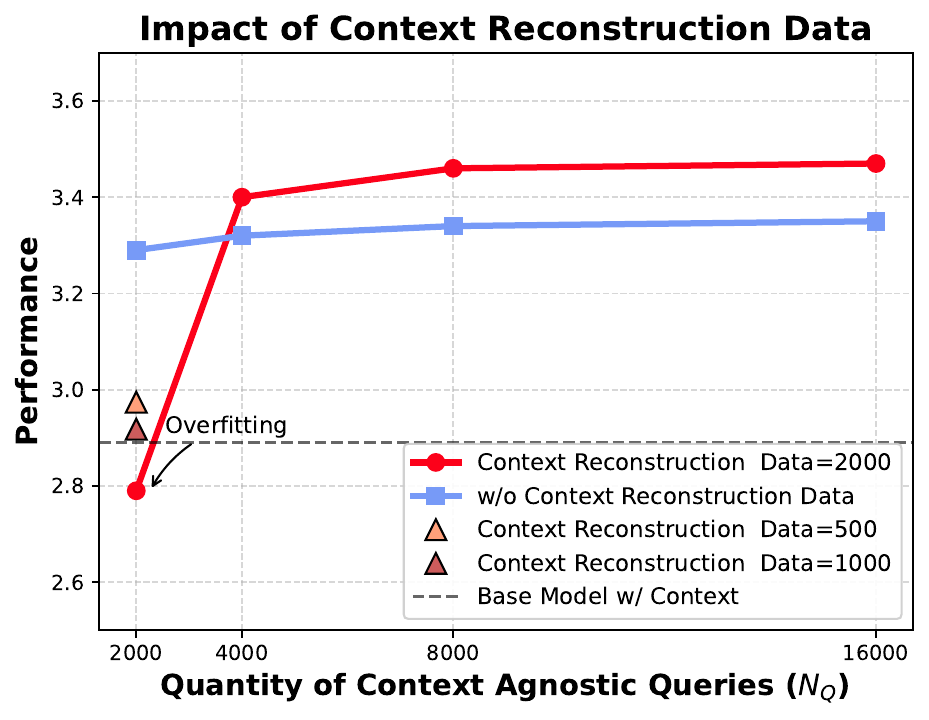}
    \caption{Training Data Ablation on CoQA.
Comparison of different Repeat Data quantities (0, 500, 1000, 2000) under varying Regularization strengths ($N_{Q}$).
}
    \label{fig:repeat_uq_scaling}
\end{figure}

\subsection{Data Strategy Analysis: The Synergy of Memory and Regularization} 
\label{sec:data_ablation}

We analyze the distinct roles and interaction effects of our two training tasks: Context Reconstruction Task and Context-Agnostic Regularization Task, as visualized in Figure \ref{fig:uq_scaling} and \ref{fig:repeat_uq_scaling}.

\paragraph{The Necessity of Regularization: Preventing Collapse.} Figure \ref{fig:uq_scaling} illustrates the impact of scaling $Q_{rnd}$ while keeping reconstruction data fixed. In the absence of regularization ($N_Q=0$), performance collapses to near zero ($0.0$), falling below even the ``No Context" baseline. This indicates that without manifold constraints, the model overfits aggressively to the reconstruction objective, degenerating into a ``parrot" that can repeat text verbatim but loses the reasoning heads required to interpret instructions.However, introducing just 1,000 $Q_{rnd}$ samples triggers a rapid recovery (score jumps to 1.89). As $Q_{rnd}$ scales to 8,000, performance asymptotically approaches the full-context upper bound. This confirms that Context-Agnostic Queries act as a crucial ``guide rail," forcing the compressed buffer tokens to reside within the high-dimensional manifold where the model's instruction-following capabilities remain active.
\paragraph{The Trade-off between Memorization and Generalization.} Figure \ref{fig:repeat_uq_scaling} reveals a critical dependency between the two tasks.We observe that in the low-regularization regime ($N_Q=2000$), increasing the quantity of reconstruction data negatively correlates with downstream performance (dropping from 3.29 to 2.78). This suggests that when the regularization signal is insufficient, explicit memory supervision dominates the gradient direction, pushing the model towards a trivial solution (verbatim repetition) that degrades reasoning ability.However, in the high-regularization regime ($N_Q=8000$), this trend is reversed. Adding reconstruction data yields consistent performance gains, peaking at 3.53. This demonstrates that sufficient regularization is a prerequisite for effective context compression. Only when the instruction-following capability is robustly anchored can the model benefit from the high-fidelity information provided by the reconstruction task without suffering from overfitting.

This analysis proves that our self-aligned optimization strategy is synergistic rather than merely additive. Reconstruction supervision serves as a ``high-risk, high-reward" signal that provides high-fidelity detail retention, but its potential is only ``unlocked" when the instruction-following manifold is robustly preserved by sufficient Manifold Regularization.

\subsection{Sensitivity Analysis \& Design Choices}
\label{sec:compression_scaling}
\paragraph{Impact of Compression Ratio}

\begin{figure}
    \centering
    \includegraphics[width=1\linewidth]{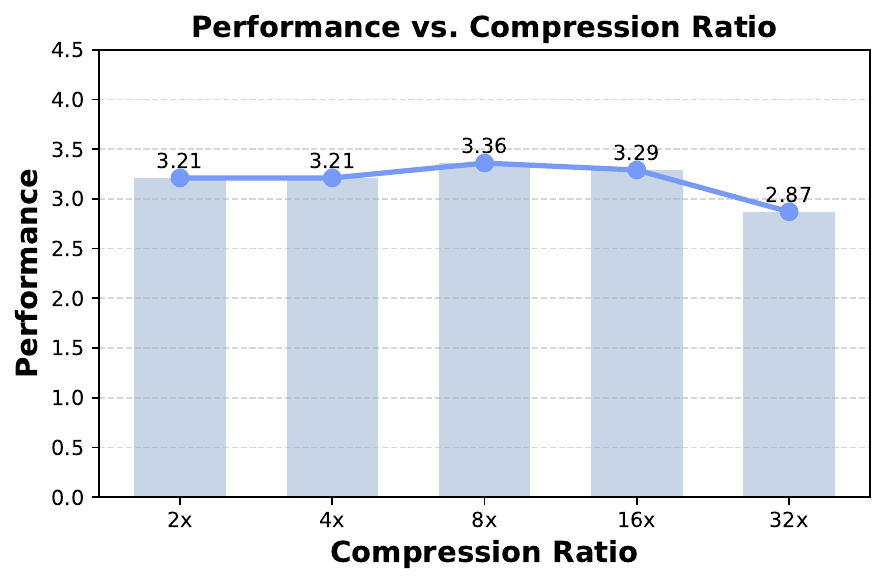}
    \caption{Impact of Compression Ratio on Model Performance. Evaluation on the CoQA dataset with compression ratios ranging from 2$\times$ to 32$\times$. }
    \label{fig:compression_scaling}
\end{figure}

We investigate the information capacity of buffer tokens by sweeping the compression ratio from 2$\times$ to 32$\times$, fixing the regularization scale at 2,000 queries. Figure \ref{fig:compression_scaling} reveals a phenomenon of Information Saturation: performance remains remarkably stable across the lower compression regimes (2$\times$ to 16$\times$), peaking at 8$\times$ (3.36). This suggests that the core semantic information of the context can be losslessly distilled into a relatively small latent space; allocating excessive tokens (2$\times$, 4$\times$) yields negligible gains, merely introducing latent redundancy.

Consequently, we adopt 16$\times$ as the strategic operating point. While there is a minor performance delta compared to the 8$\times$ peak (3.29 vs. 3.36), this is outweighed by the substantial 50\% reduction in memory footprint, positioning 16$\times$ at the optimal Pareto frontier of deployment efficiency and retention fidelity. However, we observe a distinct Channel Capacity Limit at 32$\times$, where performance sharply declines to 2.87. This inflection point signals that the bottleneck has become too restrictive, making it theoretically impossible to encode fine-grained narrative details without lossy compression.

\paragraph{Discussion of Distillation Objective (KL vs. MSE)}

\begin{figure}
    \centering
    \includegraphics[width=0.95\linewidth]{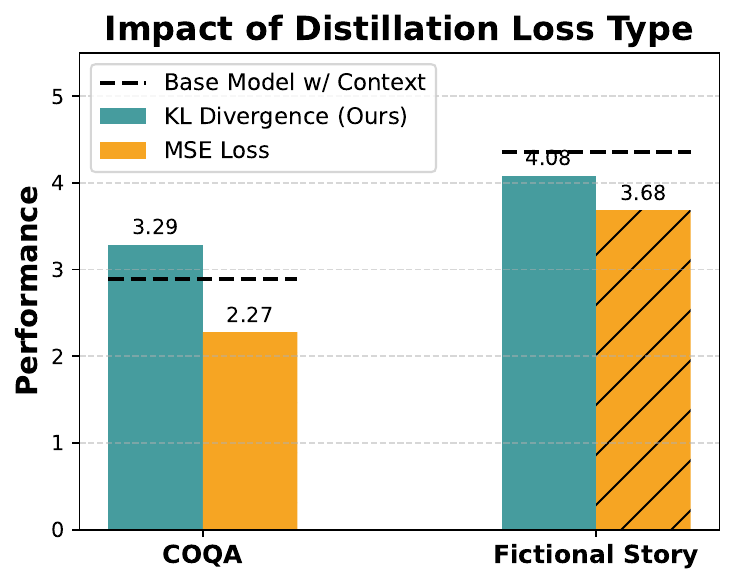}
    \caption{Ablation on Distillation Loss Type. Comparison between KL Divergence and MSE Loss on CoQA and Fictional Story.}
    \label{fig:loss_type}
\end{figure}

To validate our optimization objective, we compare the default Kullback-Leibler (KL) divergence against standard Mean Squared Error (MSE) applied to logits. Figure \ref{fig:loss_type} demonstrates the unequivocal superiority of probabilistic distillation.

On the Fictional Story dataset, the KL-trained model achieves a score of 4.08, whereas the MSE variant lags significantly at 2.87. This disparity stems from the nature of the supervision signal: MSE rigidly forces point-wise regression on logits, often over-penalizing numerical deviations that do not affect token ranking. In contrast, KL divergence aligns the entire probability distribution, allowing the buffer tokens to capture the ``Dark Knowledge"—the rich, subtle semantic relationships encoded in the teacher's soft targets. Notably, on CoQA, the KL objective enables the compressed model to marginally outperform the teacher (3.29 vs. 2.89 Base Model). We attribute this to the noise-filtering property of KL minimization combined with manifold regularization, which produces a more robust retrieval signal than the raw, noisy context.

\section{Conclusion}
In this work, we presented Latent Context Compilation, a framework that reconciles instance-specific adaptation with deployment efficiency by distilling long contexts into portable buffer tokens via a disposable LoRA. Our proposed self-aligned optimization strategy enables train data label-free optimization, successfully balancing high-fidelity context reconstruction with the preservation of general reasoning capabilities. Extensive experiments with Llama-3.1-8B demonstrate our method achieves a 16$\times$ compression ratio, significantly outperforming extractive and weight-based baselines on retrieval and summarization tasks while avoiding catastrophic forgetting. Ultimately, our method establishes a robust, plug-and-play paradigm for efficient long-context inference, offering a scalable path toward high-fidelity memory without architectural modifications.

\newpage



\bibliography{example_paper}
\bibliographystyle{icml2026}

\newpage
\appendix
\onecolumn

\section{Implementation Details}
\paragraph{Architecture \& Initialization.} 
We employ Llama-3.1-8B-Instruct as the backbone, utilizing BFloat16 precision to ensure numerical stability. The compression mechanism is implemented via a disposable LoRA module with rank $r=8$ and scaling factor $\alpha=16$. The number of buffer tokens is dynamically calculated to strictly enforce a 16$\times$ compression ratio. To facilitate rapid alignment, we initialize these buffer embeddings to the mean of the pre-trained vocabulary rather than using random initialization.
\paragraph{Optimization \& Environment.} We optimize the model for 45 epochs using the AdamW optimizer with a learning rate of $2 \times 10^{-5}$ and a linear decay schedule. As validated in our ablation studies, we utilize a KL-divergence loss to distill the teacher's full probability distribution into the buffer tokens. Experiments are conducted on 8$\times$ NVIDIA H100 (80GB) GPUs using DeepSpeed ZeRO-3, with a global effective batch size of 64 (per-device batch size of 1 with gradient accumulation steps of 8).

\section{Dataset Construction \& Prompt Templates}
\label{app:dataset_construction}

In this section, we detail the exact data construction pipeline used to generate the synthetic training set $\mathcal{D}_{surrogate}$ for test-time optimization. The process involves two key components: the mixing strategy and the specific prompt templates aligned with the Llama-3 chat format.
\paragraph{Data Mixture Strategy.}As implemented in our training script, we construct the training batch by interleaving two distinct data streams. The final training set is composed as follows:\begin{itemize}\item \textbf{Stream A: Manifold Regularization (Context-Agnostic).} We only sample queries from the Alpaca dataset and we do not the ground truth answer. 
\item \textbf{Stream B: Context Reconstruction (Memory).} We synthetically build reconstruction samples. For these samples, the user input is strictly fixed to the trigger phrase "Please repeat the context," and the target output is the verbatim raw context.\end{itemize}
\paragraph{Prompt Templates.}To ensure compatibility with the instruction-tuned backbone (\texttt{Llama-3.1-8B-Instruct}), we rigorously follow the official chat template structure. The input sequence $x$ is formatted by concatenating the system prompt, the long context (Instruction), the buffer tokens, and the random query (Input). Formally, for a given instance containing an instruction $I$ (the long context), a query $Q$, and a placeholder for buffer tokens (represented as \texttt{<GIST>} $\times k$, $k = len(I) // 16$ ), the formatted prompt is constructed as:
\begin{tcolorbox}[
    colback=blue!5,
    colframe=blue!50,
    title=Llama-3 Training Prompt Template,
    fonttitle=\bfseries,
    fontupper=\small\ttfamily
]
<|begin\_of\_text|><|start\_header\_id|>system<|end\_header\_id|>\\
\\
You are a helpful assistant.<|eot\_id|><|start\_header\_id|>user<|end\_header\_id|>\\
\\
\{Instruction\} <Gist>...<Gist> \{Input\} <|eot\_id|>\\
\\
<|start\_header\_id|>assistant<|end\_header\_id|>
\end{tcolorbox}
Where:
\begin{itemize}
\item \texttt{{Instruction}} corresponds to the raw long context (e.g., the book chapter or document).
\item \texttt{{<Gist>...<Gist>}} denotes the sequence of learnable buffer tokens inserted immediately after the context.
\item \texttt{{Input}} varies by task type:
\begin{itemize}
\item For \textbf{Regularization}: $Input \leftarrow$ "What is the capital of France?" (Random Alpaca Query)
\item For \textbf{Reconstruction}: $Input \leftarrow$ "Please repeat the context."
\end{itemize}
\end{itemize}
During optimization, the model is trained to predict the subsequent tokens (the Assistant's response) given this structured prompt.

\section{Benchmark Case Studies \& Evaluation Design}
\label{app:benchmark_cases}
To comprehensively assess the fidelity of our compressed context, we employ a dual-benchmark strategy. We utilize CoQA to evaluate conversational state tracking and coreference resolution, while our custom Synthetic Fictional Story benchmark probes the model's capacity for fine-grained semantic interpretation across varying difficulty levels.\subsection{CoQA: Conversational State Tracking}The CoQA (Conversational Question Answering) dataset challenges the model to maintain a coherent memory state across multi-turn dialogues. Unlike standard single-turn QA, CoQA heavily relies on coreference resolution (e.g., resolving pronouns like "he", "it", or "that" to entities mentioned in previous turns).
\begin{tcolorbox}[colback=gray!5,colframe=gray!50,title=Case Study from CoQA: The Story of Asta, fontupper=\small]\textbf{Context Snippet:} ``...One day, a bottle floated by over the heads of Asta and his friends. They looked up and saw the bottle. What is it?' said Asta's friend Sharkie. It looks like a bird's belly,' said Asta..."\vspace{0.2cm}
\\
\textbf{Evaluation Questions:}
\begin{itemize}\item \textit{Q:} "What was the name of the fish?" $\rightarrow$ \textit{A:} "Asta"
\item \textit{Q:} "What looked like a bird's belly?" $\rightarrow$ \textit{A:} "A bottle"

\item \textit{Q:} "Who said {that}?" 
\textit{A:} "Asta"

\item \textit{Q:} "Was Sharkie a friend?" $\rightarrow$ \textit{A:} "Yes"
\end{itemize}\end{tcolorbox}

\subsection{Synthetic Fictional Story: Stratified Reasoning}
Standard benchmarks often lack the granularity to distinguish between "keyword matching" and "true understanding." To address this, we constructed a Fictional Story dataset where each narrative is paired with questions stratified by difficulty: Easy (Retrieval), Medium (Inference), and Hard (Thematic Analysis).

\begin{tcolorbox}[colback=gray!5,colframe=gray!50,title=Case Study from Synthetic Fictional Story: The Baker of Eldermere, fontupper=\small]\textbf{Context Snippet:} ``A baker, Mr. Thorne, bakes bread using abstract concepts (starlight, whispers) instead of ingredients. A newcomer, Elara, discovers this secret and offers a stone (a memory) as payment, leading to the creation of a new bread... "\vspace{0.2cm}
\\
\textbf{Evaluation Questions:}
\\
\textbf{Level 1: Surface-Level Retrieval (Easy)} \\
\begin{itemize}\item \textit{Q:} "What time does the bread appear every morning?"\item \textit{A:} "8:15 AM." (Exact match required)\end{itemize}\textbf{Level 2: Cross-Sentence Inference (Medium)} \\
\begin{itemize}\item \textit{Q:} "Why was it strange that the bread appeared at 8:15?"\item \textit{A:} "Because no one ever saw Mr. Thorne bake or receive deliveries." (Requires synthesizing the "mystery" aspect).\end{itemize}\textbf{Level 3: Thematic \& Symbolic Analysis (Hard)} \\
\begin{itemize}\item \textit{Q:} "What is the symbolic meaning of Mr. Thorne's baking process?"\item \textit{A:} "He is preserving memories and feelings that would otherwise fade."\end{itemize}\end{tcolorbox}

\section{LLM-as-Judge Case Studies} 
\label{app:judge_cases}

We employ GPT-4o to score model responses on a scale of 0--5 based on accuracy and faithfulness to the source context. Below are representative examples of how specific error types are penalized.

\subsection{Case 1: Hallucination \& Fabrication (Score: 1--2)} \begin{tcolorbox}[colback=blue!5,colframe=blue!50,title=Severe Fabrication, fontupper=\small] \textbf{Question:} Were they excited? \ \\
\textbf{Ground Truth:} Unknown (Not explicitly stated).

\textbf{Model Response:} \ "They were super excited! They had never had a pen pal before and the thought of having a friend from a far away place was amazing! Asta and Sharkie quickly set to work looking for another bottle..."

\textbf{Judge's Reasoning:} \ \textbf{Score: 1.} The generated response introduces significant fabricated details—such as the specific motivation of having a pen pal and the immediate action of searching for another bottle—which are entirely absent from the context. While the sentiment might be inferred, the factual hallucination warrants a penalty. \end{tcolorbox}

\subsection{Case 2: Partial Accuracy / Minor Error (Score: 3)} \begin{tcolorbox}[colback=blue!5,colframe=blue!50,title=Quantity Mismatch, fontupper=\small] \textbf{Question:} What did Bolt finally create, and can you describe it? \ \\
\textbf{Ground Truth:} He finally built {two} magnificent silver wings.

\textbf{Model Response:} \ "Bolt created a magnificent silver {wing}."

\textbf{Judge's Reasoning:} \ \textbf{Score: 3.} The response partially captures the core event (creating silver wings) but contains a factual inaccuracy regarding quantity ("a wing" vs. "two wings"). This renders the answer incomplete and slightly misleading, preventing a higher score. \end{tcolorbox}

\subsection{Case 3: Correct but Verbose (Score: 4--5)} \begin{tcolorbox}[colback=blue!5,colframe=blue!50,title=Correct with Extraneous Reasoning, fontupper=\small] \textbf{Question:} Who could read the note? \ \\\textbf{Ground Truth:} Asta's papa.

\textbf{Model Response:} \ "Asta's papa could read the note because he is Asta's parent, and parents are usually able to read."

\textbf{Judge's Reasoning:} \ \textbf{Score: 4.} The response correctly identifies the key entity ("Asta's papa") required by the question. Although it adds external reasoning ("parents are usually able to read") that is not in the text, this addition does not contradict the context or harm the factual accuracy of the answer. \end{tcolorbox}

\subsection{Case 4: Perfect Instruction Following (Score: 5)} \begin{tcolorbox}[colback=blue!5,colframe=blue!50,title=Exact Retrieval, fontupper=\small] \textbf{Question:} What is the name of the little robot in the story? Use only 1 word. \ \\
\textbf{Ground Truth:} Bolt.

\textbf{Model Response:} \ "Bolt"

\textbf{Judge's Reasoning:} \ \textbf{Score: 5.} The response is perfect. It correctly retrieves the specific entity ("Bolt") requested by the prompt and strictly adheres to the formatting constraint ("Use only 1 word"). \end{tcolorbox}

\section{Computational Efficiency \& Real-World Deployment}
\label{app:efficiency_applications}
A common critique of instance-specific optimization is the latency introduced by the training phase. In this section, we analyze the amortization dynamics of Latent Context Compilation and discuss specific deployment scenarios where our "Write-Once, Read-Many" paradigm offers a decisive advantage over standard RAG or long-context inference.
\subsection{Computational Efficiency }
\textbf{One-Time Cost vs. Recurring Savings.} The compilation process incurs an initial, fixed computational overhead ($T_{compile}$) to optimize the disposable LoRA. However, this is a one-time sunk cost. Once compiled, the heavy raw context $C$ is permanently replaced by the compact buffer tokens $T_{buf}$ (where $|T_{buf}| \ll |C|$). Consequently, for all subsequent queries $Q_1, ..., Q_N$, the inference latency is reduced from linear scaling with the full context $O(|C| + |Q|)$ to a minimal constant overhead $O(|T_{buf}| + |Q|)$.

\subsection{Potential Deployment Scenarios}Leveraging the portability of our buffer tokens, Latent Context Compilation unlocks high-value applications that are currently infeasible for standard long-context models:

\textbf{1. Server-Side Personalized Agents (Long-Term Memory).} For personal AI assistants, a user's history can span months of conversation. Standard methods either truncate this history (losing key details) or re-process the entire sequence every turn (prohibitively expensive). Our framework allows the agent to maintain a "Running Buffer" of the user's persona and past interactions. The buffer can be updated periodically (e.g., nightly compilation), allowing for infinite memory retention with constant-time inference latency, effectively decoupling memory depth from serving cost.

\textbf{2. Enterprise Knowledge Bases (Global Reasoning).} In scenarios like analyzing a 100-page merger agreement or a fiscal year report, professionals often ask dozens of exploratory questions. RAG systems often fail here due to "fragmentation"—missing the cross-document relationships. By compiling the \textit{entire} document into a holistic latent state, our method enables the model to answer global reasoning questions (e.g., "How do the liability clauses interact with the payment schedule?") that retrieval-based methods struggle with, without the recurring cost of processing 100 pages for every question.

\textbf{3. On-Device Intelligence.} Deploying long-context capabilities on consumer hardware (e.g., smartphones, laptops) is strictly constrained by VRAM and bandwidth. Standard KV caching for a 100k-token context can easily consume tens of gigabytes of memory, exceeding the capacity of most edge devices. Latent Context Compilation solves this by compressing the context into a lightweight set of buffer tokens (e.g., < 1MB). This allows users to "compile" sensitive documents locally (or fetch pre-compiled buffers) and perform complex reasoning entirely offline. This "Local-First" architecture not only eliminates cloud inference costs but also guarantees data privacy, as the raw context never leaves the device during the query phase.





\end{document}